\RequirePackage[OT1]{fontenc}
\documentclass[letterpaper, 10 pt, conference]{ieeeconf}  %

\IEEEoverridecommandlockouts                              %

\usepackage{adjustbox}
\usepackage{makecell}
\usepackage{multirow}
\usepackage{amsmath}
\usepackage{graphicx}
\usepackage{siunitx}
\usepackage{booktabs}
\usepackage{subcaption}
\usepackage{amsmath}
\usepackage{amssymb}
\usepackage{xparse}
\usepackage{comment}
\usepackage{hyperref}
\usepackage[capitalize]{cleveref}
\usepackage[style=ieee,autocite=footnote,maxcitenames=1,minnames=1,maxalphanames=4,minalphanames=3,natbib=true,url=false, doi=false]{biblatex}
\DefineBibliographyStrings{english}{%
    andothers = {\normalfont et\addabbrvspace al\adddot}
}

\usepackage{tikz}
\usetikzlibrary{positioning, arrows.meta, shapes.geometric, backgrounds, calc}

\usetikzlibrary{fit}
\definecolor{oiBlue}{RGB}{0,114,178}
\tikzset{
    newnode/.style={
            draw=oiBlue,
            text=oiBlue,
            fill=oiBlue!8,   %
            rounded corners,
            minimum width=2.3cm,
            minimum height=1cm,
            text width=2.3cm,
            align=center
        },
    newout/.style={
            draw=oiBlue,
            text=oiBlue,
            fill=oiBlue!8,
            rounded corners,
            minimum width=2cm,
            minimum height=1cm,
            align=center
        },
    newarrow/.style={-latex, thick, draw=oiBlue},
    newlineblue/.style={-, thick, draw=oiBlue},
    blacknode/.style={
        draw, rectangle, minimum width=2.3cm, minimum height=1cm, text width=2.3cm, align=center, rounded corners
    }
}

\title{Unpaired RGB-Thermal Gaussian-Splatting Using Visual Geometric Transformers}

\author{Jean Cordonnier, Chenghao Xu, Olga Fink, and Malcolm Mielle%
\thanks{*This work was supported by INNOSUISSE grant 105.237 IP-ICT}%
\thanks{Jean Cordonnier, Chenghao Xu, and Olga Fink are with Ecole Polytechnique Federale de Lausanne, Lausanne, Switzerland.
            {\tt\small
                @epfl.ch}}%
    \thanks{Malcolm Mielle is with Schindler EPFL Lab, Lausanne, Switzerland
            {\tt\small malcolm.mielle@schindler.com}}%
}

\begin{document}

\maketitle
\thispagestyle{empty}
\pagestyle{empty}

\begin{abstract}

    Multi-modal novel view synthesis (NVS) combining RGB and thermal imagery enables precise 3D scene reconstruction with visual and thermal information.
    However, existing methods typically rely on precisely calibrated RGB-thermal image pairs or stereo setups, limiting scalability and practical deployment.
    To address this, we introduce a framework for unpaired RGB-thermal NVS that leverages VGGT, a 3D feed-forward transformer architecture, to independently estimate camera poses for each modality.
    The pose sets are then aligned using the Procrustes algorithm with a cross-modal feature matcher, enabling joint registration without paired calibration.
    Building on this alignment, we further propose a multi-modal 3D Gaussian Splatting approach that learns directly from unpaired RGB and thermal images.
    Experiments on diverse scenes demonstrate that our method achieves competitive performance in thermal view synthesis while maintaining RGB fidelity.
    Moreover, we show that existing reconstruction approaches can produce modality-specific reconstructions that lack cross-modal consistency.
    We thus introduce a benchmarking framework to rigorously evaluate both per-modality image synthesis and the multi-modal coherence of reconstructed scenes.

\end{abstract}

\section{INTRODUCTION}

Multi-modal visual reconstruction has attracted growing attention for capturing complementary aspects of physical scenes.
For example, fusing RGB and thermal imaging---which captures heat signatures---enables applications such as SLAM in low-light conditions~\cite{10802480} and estimation of thermal weaknesses in urban infrastructure~\cite{hassanThermoNeRFMultimodalNeural2025}.

To construct unified 3D representations from RGB and thermal imagery, multi-modal novel view synthesis (NVS) methods have been proposed, including ThermoNeRF~\cite{hassanThermoNeRFMultimodalNeural2025} and Thermal Gaussian Splatting~\cite{liu2025thermalgs}.
However, those models typically requires a preprocessing step before scene reconstruction where, for each image in the input set, the camera's extrinsic and intrinsic parameters must be estimated, typically from Structure-from-Motion (SfM) pipelines~\cite{schoenberger2016sfm} .
Since thermal images are low-contrast images that capture thermal radiations that are scattered and reflected by objects and the environment, they are naturally textureless and soft---a phenomenon known as the ghosting effect~\cite{Bao:24}.
Thus, feature-matching based SfM methods struggle to converge and typically only work well on high-quality thermal dataset with abundant visual detail~\cite{hassanThermoNeRFMultimodalNeural2025}.

Hence, prior work on RGB-Thermal (RGB-T) scene reconstruction typically rely on stereo setups with paired RGB and thermal images, where the estimated RGB poses are directly used for the thermal images~\cite{hassanThermoNeRFMultimodalNeural2025,lu2025thermalgaussian,chassaingThermoxelsVoxelbasedMethod2025,10.1007/978-3-031-92805-5_6}.
However, stereo setups across different modalities are costly and difficult to calibrate compared to single-camera setups.
Those contraints limit the flexibility and scalability of multi-modal reconstruction, for example for multi robot mapping scenarios where each robot carries different, but complementary, sensors;
There is a need for cross-modal reconstruction strategies where RGB and thermal images can be acquired independently before reconstructing a multi-modal scene.

To tackle this issue, we propose a framework for RGB-T scene reconstruction from two unpaired sets of RGB and thermal images.
We first show how VGGT, an RGB 3D feed-forward transformer architecture, can create unimodal reconstructions for both RGB and thermal image sets without the need for fine-tuning on thermal data.
Then, we estimate the RGB and thermal camera parameters independently using VGGT and propose a strategy to align both sets of poses using correspondences obtained with a state-of-the-art RGB-to-Thermal feature matcher.
Finally, we propose a Gaussian Splatting training strategy that leverages the set of aligned (but unpaired) RGB and thermal images to reconstruct the scene.
We evaluate our method on nine RGB-T scenes, demonstrating competitive performance in thermal view synthesis while maintaining RGB fidelity under unpaired settings.
Furthermore, we show that existing reconstruction approaches can produce modality-specific reconstructions that lack cross-modal consistency, and we introduce a benchmarking framework to rigorously evaluate both per-modality image synthesis and the multi-modal coherence of reconstructed scenes.

\section{RELATED WORK}

\begin{figure*}[t]
    \centering
    \vspace{1mm}
    \input{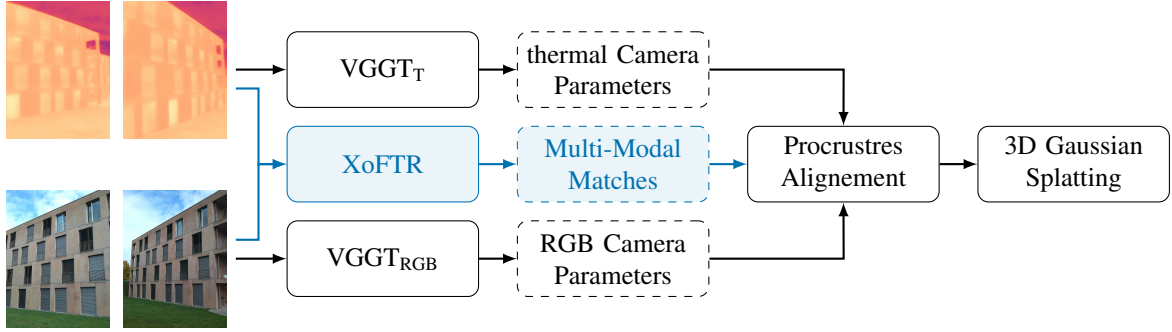}
    \caption{
        Flowchart of the method's steps.
        Solid boxes denote processing steps, while dashed boxes indicate outputs.
        Given two sets of images from a scene (RGB and thermal), VGGT is first used to determine the camera parameters of each set independently.
        Then, features between RGB and thermal images are found using XoFTR; those features are used to align both sets of camera parameters, in translation, rotation, and scale.
        Finally, given the uniform set of camera parameters, a representation of the scene is generated using 3D Gaussian Splatting.
    }
    \label{fig:method:framework}
    \vspace{-5mm}
\end{figure*}

\subsection{Transformer for 3D Geometric Reconstruction}
\label{sec:relatedwork:transformer}

Structure-from-Motion (SfM) reconstructs 3D scene geometry and camera parameters given a set of images.
SfM approaches, such as COLMAP~\cite{schoenberger2016sfm}, usually decompose this problem into several steps: keypoint extraction and matching, relative pose estimation, and incremental reconstruction with triangulation and bundle adjustment.
Recent advances in deep learning have challenged this approach.
For instance, DUSt3R~\cite{wang2024dust3r} showed that a single forward pass of a transformer architecture can directly regress the scene geometry and camera parameters.
To scale this approach, \textcite{duisterhof2024mast3r} aligned local DUSt3R reconstruction into a global coordinate system.
However, this process remains computationally expensive, especially the optimization stage~\cite{wang2024dust3r}.

More recently, VGGT~\cite{wang2025vggt} presented a transformer architecture with alternating attention that can process multiple input images and directly infer all 3D attributes of a scene, including extrinsic and intrinsic camera parameters, point maps, depth maps, and 3D point tracks, within seconds.
Notably, VGGT has been reported to outperform COLMAP in both 3D initialization and camera pose estimation~\cite{wang2025vggt}.
Transformer-based 3D reconstruction methods have shown strong adaptability and the ability to handle complex scenes, but their application in multi-modal settings such as RGB–thermal reconstruction remains largely unexplored.

\subsection{Multi-modal 3D Reconstruction and Novel View Synthesis}

Traditional 3D reconstruction methods, such as SfM, depend on dense, feature-rich input data.
In contrast, Neural Radiance Field (NeRF)~\cite{mildenhall2021nerf} introduced implicit 3D scene representations learned from sparse RGB images, enabling high-quality novel-view synthesis.
However, while recent NeRF models are more lightweight and efficient~\cite{sun20243QFP:_}, computing novel views remains computationally expensive.
3D Gaussian Splatting  (3DGS)~\cite{kerbl20233d}
represents a scene using explicit 3D Gaussians parameterized by position $\mu$, covariance $\Sigma$, opacity $\alpha$, and color $c$. %
Such explicit representation supports real-time rendering through fast rasterization, making 3DGS a compelling alternative to NeRF~\cite{sunHighFidelitySLAMUsing2024}.%

Both NeRF and 3DGS have been extended to multimodal settings, incorporating thermal images~\cite{hassanThermoNeRFMultimodalNeural2025,chassaingThermoxelsVoxelbasedMethod2025}, semantic information~\cite{xu2025Exploi}, or hyper-spectral images~\cite{Thirgood_2025_CVPR}, thereby enabling diverse applications, such as building information extraction~\cite{xu2025Exploi}.
However, both unimodal and multimodal approaches still depend on external frameworks (e.g., SfM) to estimate camera poses and intrinsics from input images.
For instance, \textcite{hassanThermoNeRFMultimodalNeural2025}, \textcite{chassaingThermoxelsVoxelbasedMethod2025}, and \textcite{lu2025thermalgaussian} assume paired RGB-thermal inputs and use COLMAP to recover RGB poses, which are then transferred to the thermal images.
This assumption limits real-world deployment, where modality-specific inputs are often collected separately at different times, by different users, or across multiple robotic platforms.
In contrast, our work decouple the RGB and thermal image inputs under unpaired settings, allowing both modalities to be sampled from different views while enabling consistent multi-modal reconstruction.

\section{Method}
\label{sec:method}

We present a method for unpaired RGB-Thermal (RGB-T) novel view synthesis (NVS) that performs independent pose estimation followed by cross-modal alignment.
Given two unpaired sets of images (i.e., RGB and thermal) of the same scene, we first independently estimate their camera poses using the VGGT architecture.
Next, we introduce a cross-modal feature matching strategy to establish correspondences between RGB and thermal images.
These correspondences are used to compute a similarity transformation, accounting for translation, rotation, and scale, that aligns the intrinsic and extrinsic camera parameters across modalities.
Finally, using the aligned camera poses, we train a 3DGS thermal model to achieve multi-modal NVS.
A representation of our framework is shown in \cref{fig:method:framework}.

\subsection{Decoupled VGGT initialization}
\label{sec:method:vggt}

\begin{figure*}[t]
    \vspace{2mm}
    \centering
    \begin{subfigure}[t]{0.24\textwidth}
        \centering
        \includegraphics[width=\textwidth]{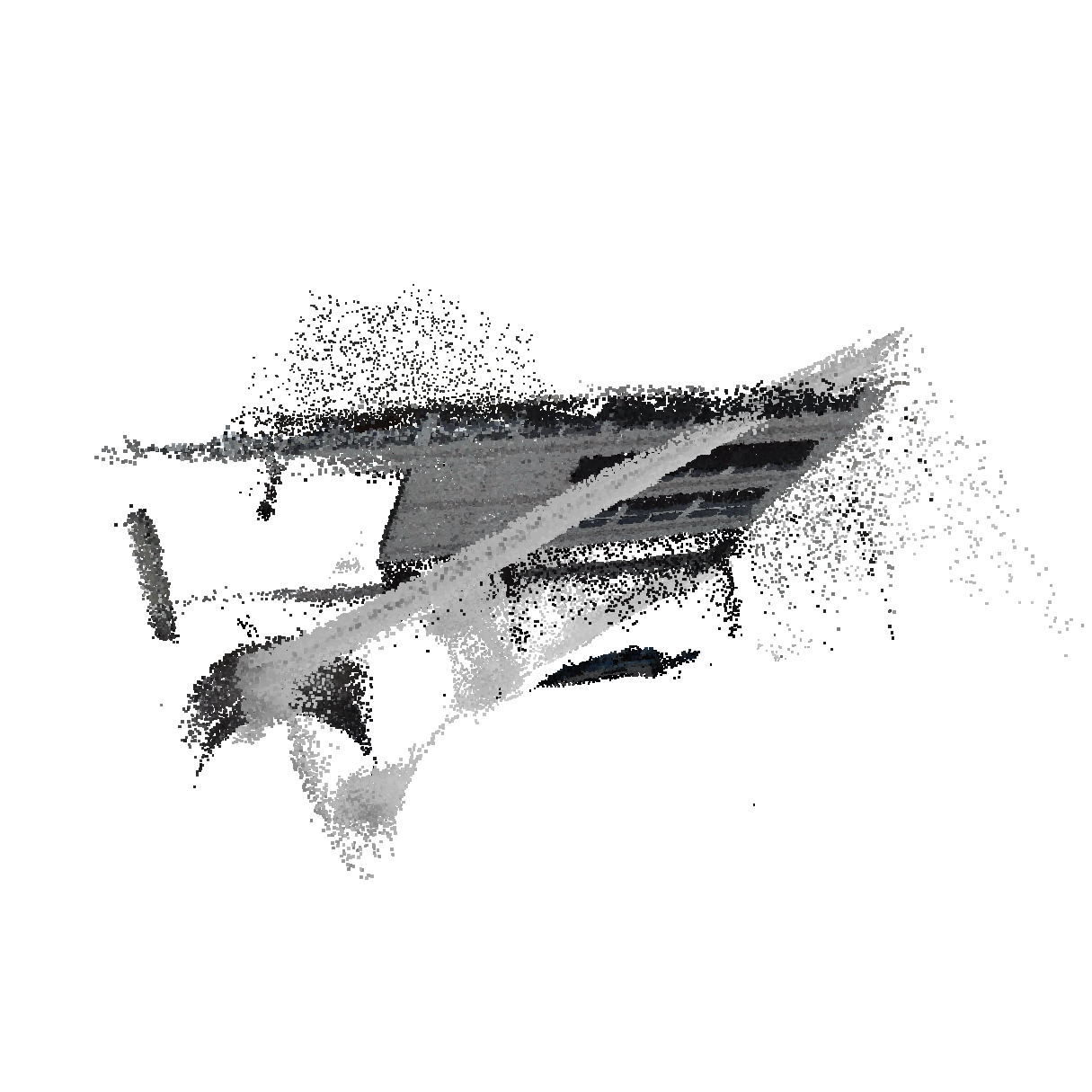}
    \end{subfigure}
    \hfil
    \begin{subfigure}[t]{0.24\textwidth}
        \centering
        \includegraphics[width=\textwidth]{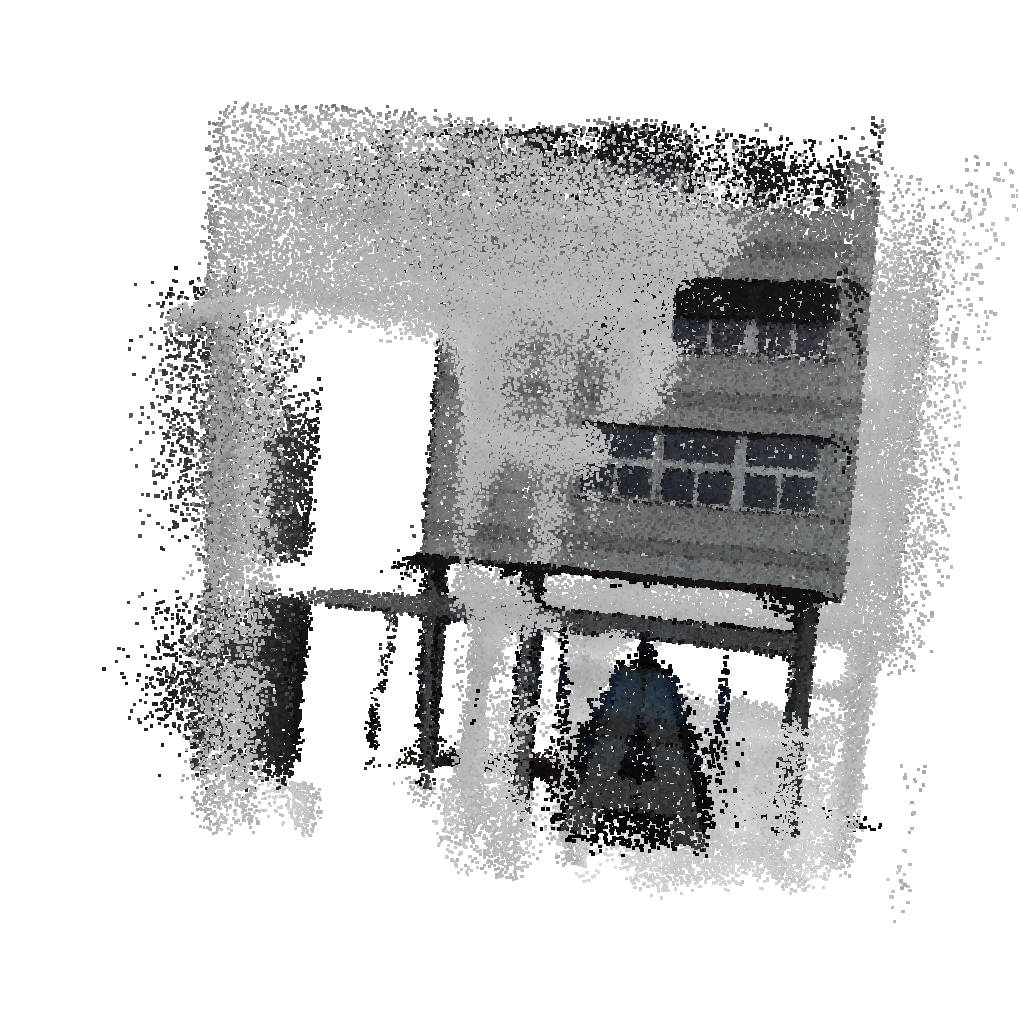}
    \end{subfigure}
    \hfil
    \begin{subfigure}[t]{0.24\textwidth}
        \centering
        \includegraphics[width=\textwidth]{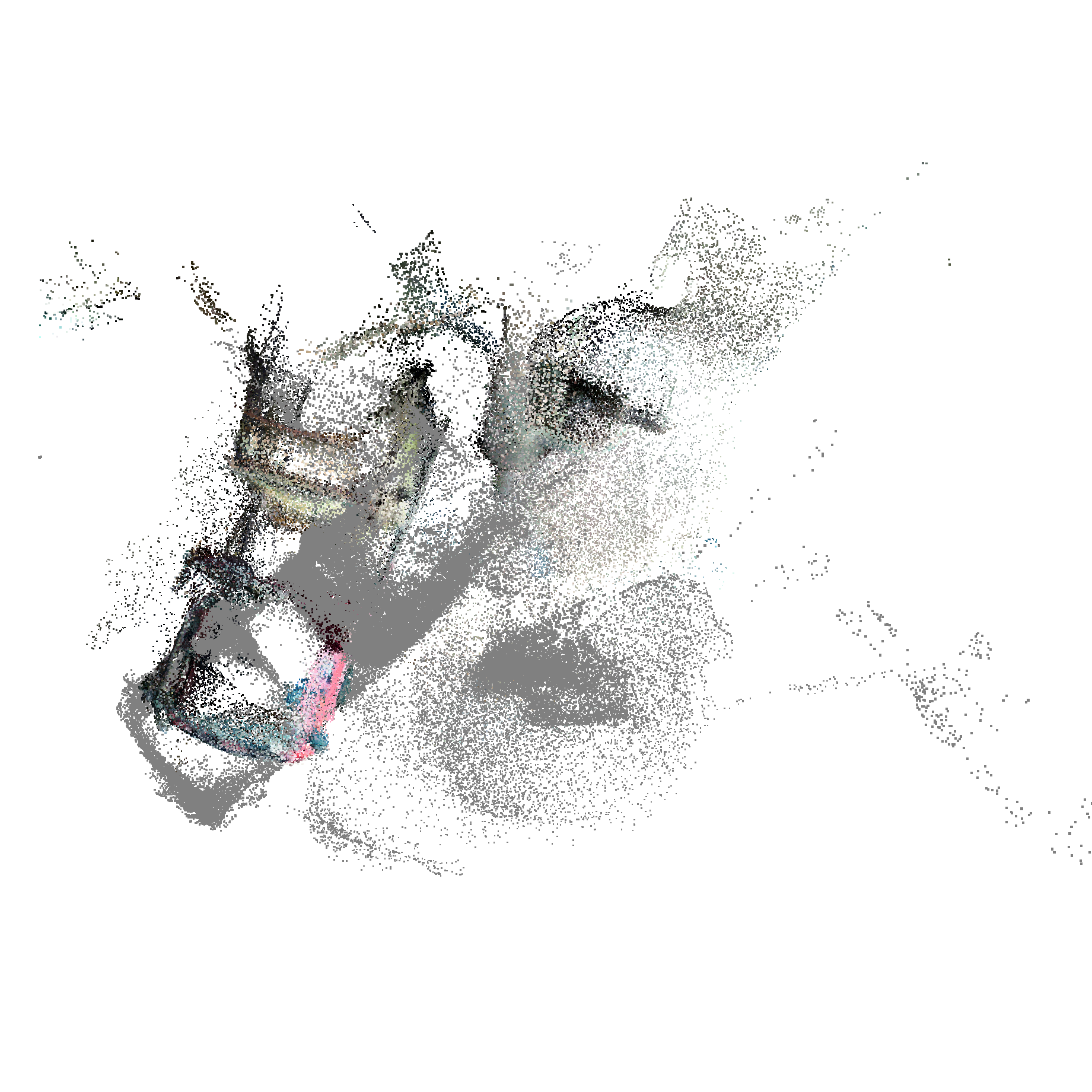}
    \end{subfigure}
    \hfil
    \begin{subfigure}[t]{0.24\textwidth}
        \centering
        \includegraphics[width=\textwidth]{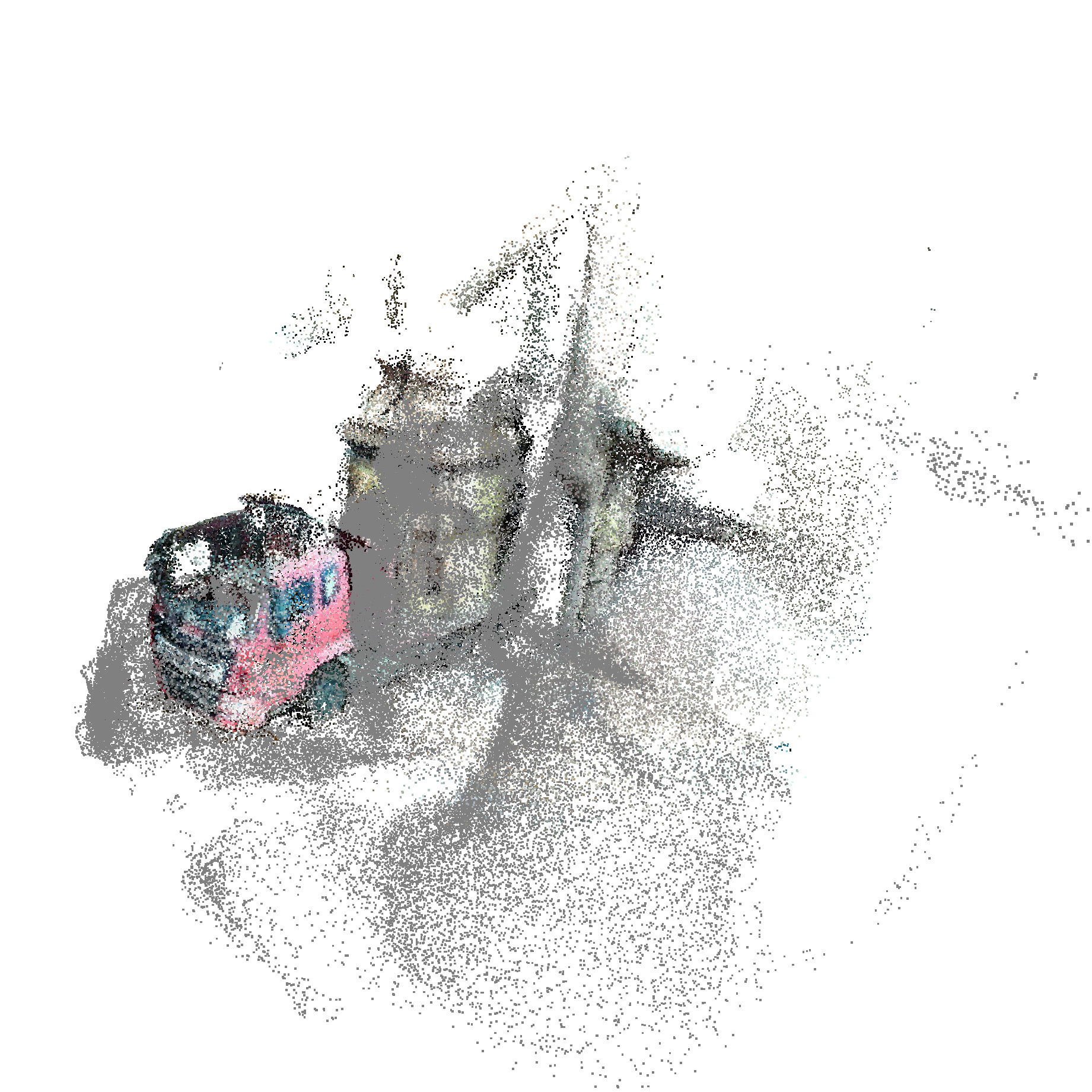}
    \end{subfigure}
    \caption{
        Thermal and RGB scene reconstructions are obtained for VGGT without fine-tuning on each modality.
        While the scene is accurately reconstructed in both modalities, there is no guarantee of alignment and scale between them.
    }
    \label{fig:method:vggttir}
    \vspace{-5mm}
\end{figure*}

As presented in \cref{sec:relatedwork:transformer}, VGGT is a feed-forward neural network that infers 3D scene attributes---including extrinsic and intrinsic camera parameters, point maps, depth maps, and 3D point tracks---from a set of sparse RGB images.
While VGGT has demonstrated strong performance on RGB scenes, its applicability to other modalities, such as thermal imagery, remains unexplored.

In this paper, we hypothesize that VGGT's robustness to scene variations (e.g., illumination or reflections) enables it to infer accurate 3D attributes from thermal images \emph{without fine-tuning}.
However, thermal images are single-channel representing per-pixel temperature and VGGT takes three-channel RGB images as input.
Thus, we replicate the thermal channel across all three input channels and these thermal images are then processed by VGGT to extract scene attributes.
Remarkably, even though noisier than their RGB counterparts, VGGT can directly reconstructs thermal scenes without finetuning, even when applied to this previously unseen modality (see \cref{fig:method:vggttir}).
This highlights the adaptability and generalization capacity of 3D feed-forward transformers.

However, while the RGB and thermal scenes reconstructed by VGGT depict the same scene, there is no guarantee of spatial alignment between the two, resulting in two sets of camera poses with arbitrary differences in position, rotation, and scale.

\subsection{Procrustres Alignment}
\label{sec:method:procrustes}

To align the camera poses of the RGB and thermal datasets, we use XoFTR~\cite{tuzcuouglu2024xoftr} to establish correspondences between thermal and RGB images.
Outliers are filtered using MAGSAC~\cite{barath2019magsac}, and only image pairs with more than $t$ correspondences are kept for the next step.

Next, we refine the correspondences using VGGT scene reconstruction, discarding those linked to low-confidence 3D points of the reconstruction.
Specifically, we reconstruct a point cloud for each modality ($\mathbf{P}_{\text{RGB}}$ and $\mathbf{P}_{\text{T}}$) using the estimated camera parameters and depth maps from VGGT outputs, where each point is assigned a confidence score.
Correspondences associated with point cloud points below the 50\% percentile of all confidences of their respective modality are discarded.

For each XoFTR feature, we compute its 3D projection onto the corresponding point cloud, $\mathbf{P}_{\text{RGB}}$ or $\mathbf{P}_{\text{T}}$, thereby obtaining two modality-specific 3D correspondences as
\begin{equation}
    \text{PF}_\text{RGB} = \{\mathbf{p}_{i,\text{RGB}} \in \mathbb{R}^3\}_{i=1}^N
\end{equation}
\begin{equation}
    \text{PF}_\text{T} = \{\mathbf{p}_{i,\text{T}} \in \mathbb{R}^3\}_{i=1}^N
\end{equation}
where $\mathbf{p}_{i,T}$ denotes a point in the thermal reconstructed point cloud and $\mathbf{p}_{i,RGB}$ the associated point in the RGB reconstructed point cloud.

To align the two point clouds from each modality, we adapt the trimmed ICP algorithm \textcite{chetverikov2002trimmed} by replacing the ICP step with a Procrustes $\text{Sim}(3)$ alignment.
Procrustes estimates the similarity transform $(s,\mathbf{R},\mathbf{t})$ such that
\begin{equation}
    \begin{split}
         & \min \sum_{i=1}^N \big\| \mathbf{p}_{i,\text{RGB}} - (s\,\mathbf{R}\mathbf{p}_{i,\text{T}} + \mathbf{t}) \big\|^2 \\
         & \text{with} \quad s>0,\,\mathbf{R}\in SO(3),\,\mathbf{t}\in\mathbb{R}^3 .
    \end{split}
\end{equation}
The closed-form solution is found via singular value decomposition (SVD) of the cross-covariance between the two point sets, yielding the optimal rotation $\mathbf{R}$, scale $s$, and translation $\mathbf{t}$.
After each step, 10\% of the points with the biggest residuals are discarded and the transformation is recomputed without them.

Once the similarity transform $(s,\mathbf{R},\mathbf{t})$ has been estimated,
It can be written as a $4\times 4$ homogeneous matrix.
\begin{equation}
    \mathbf{T}_{Sim(3)} \;=\;
    \begin{bmatrix}
        s\,\mathbf{R}   & \mathbf{t} \\
        \mathbf{0}^\top & 1
    \end{bmatrix}.
\end{equation}

Each thermal camera pose from VGGT is given by a homogeneous extrinsic matrix
\begin{equation}
    \mathbf{T}_\text{T} \;=\;
    \begin{bmatrix}
        \mathbf{R}_c    & \mathbf{t}_c \\
        \mathbf{0}^\top & 1
    \end{bmatrix}
\end{equation}
which can be reprojected into the RGB frame using
\begin{equation}
    \mathbf{T}_\text{T}^{*} \;=\; \mathbf{T}_\text{T}\,\mathbf{T}_{Sim(3)}^{-1} \, .
\end{equation}

Thus, the corresponding camera poses from each modality, $\mathbf{T}_\text{T}^{*}$ and $\mathbf{T}_\text{RGB}$, are expressed in the same reference frame under the transformation $(s,\mathbf{R},\mathbf{t})$ and
can be directly used as input to multi-modal NVS models.

It should be noted that for two scenes (Building A Spring and DimSum), we also had to use a RANSAC-based~\cite{fischler1981random} outlier rejection before applying trimmed-Procustes.

\begin{table*}[t]
  \centering
  \vspace{1mm}
  \caption{
    Given that Thermal Gaussian (T-GS) corresponds to an ideal situation where the RGB and thermal images are perfectly aligned, our method obtained, from unpaired image, competitive results.
  }
  \label{tab:per-scene-metrics-all}
  \begin{tabular}{lcccccc}
    \toprule
    Method              & \multicolumn{3}{c}{RGB} & \multicolumn{3}{c}{Thermal}                                                                                                          \\
    \cmidrule(lr){2-4} \cmidrule(lr){5-7}
                        & PSNR [dB] $\uparrow$    & SSIM $[-]$ $\uparrow$       & LPIPS $[-]$ $\downarrow$ & PSNR [dB] $\uparrow$ & SSIM $[-]$ $\uparrow$ & MAE [$^\circ$C] $\downarrow$ \\
    \midrule
    T-GS$_{\text{RGB}}$ & 18.149                  & 0.563                       & 0.383                    & 26.274               & 0.844                 & 2.554                        \\
    T-GS$_{\text{T}}$   & 16.487                  & 0.497                       & 0.451                    & 25.107               & 0.825                 & 2.532                        \\
    Ours$_{\text{RGB}}$ & 17.232                  & 0.535                       & 0.431                    & 21.315               & 0.771                 & 4.914                        \\
    Ours$_{\text{T}}$   & 17.615                  & 0.558                       & 0.414                    & 25.595               & 0.851                 & 2.509                        \\
    \bottomrule
  \end{tabular}
\end{table*}

\subsection{3D Thermal Gaussian Splatting}

Finally, to obtain a multimodal 3D scene reconstruction for NVS, we implement an RGB-T Gaussian Splatting framework that, unlike previous methods~\cite{lu2025thermalgaussian, hassanThermoNeRFMultimodalNeural2025}, does not need paired images as input.

Inspired by \textcite{lu2025thermalgaussian}, we also adopt a 3DGS framework where each Gaussian is represented by its position $\mu$, covariance matrix $\Sigma$, opacity $\alpha$, RGB color $c$, and thermal color $t$.
This joint representation of the space ensures that the two modalities share the same geometry, given that the Gaussians encode both RGB and thermal information, and thus have the same position, shapes, and opacity.
Note that color terms $c$ and $t$ incorporate spherical harmonics encoding, allowing the appearance of each Gaussian to be view-dependent.

To optimize the representation, we adopt the SSIM loss:
\begin{equation}
    \begin{split}
        \mathcal{L}_{\text{total}} = {} & \mathcal{L}_{\text{SSIM}}(\text{Temp}_{\text{pred}}, \text{Temp}_{\text{gt}})+ \\
                                        & \mathcal{L}_{\text{SSIM}}(\text{RGB}_{\text{pred}}, \text{RGB}_{\text{gt}})
    \end{split}
    \label{eq:full_loss}
\end{equation}
with
\begin{equation}
    \begin{split}
        \mathcal{L}_{\text{SSIM}}(X, Y) = {} & (1 - \lambda) \cdot \text{MAE}(X, Y) + \\
                                             & \lambda \cdot (1 - \text{SSIM}(X, Y))
    \end{split}
\end{equation}

It should be noted that, contrary to \textcite{lu2025thermalgaussian}, we do not use the smoothing loss for the thermal modality, as we did not observe improvements for image synthesis.

To handle unpaired camera inputs, we shuffle RGB and thermal images randomly at the start of each epoch and render a random RGB and thermal image pair (without spatial alignment constraints).
The loss is computed as in \cref{eq:full_loss}, and the error is backpropagated as if the images were paired.

\begin{table*}[t]
  \vspace{1mm}
  \caption{
    Image synthesis for the 4 scenes from Thermal Gaussians and ThermoScene.
    \textit{GT} correspond to the ground-truth images and \textit{Ours} correspond to image synthesis using our method from either the RGB or thermal camera pose.
  }
  \label{fig:eval:dataset}
  \centering
  \resizebox{0.8\textwidth}{!}{%
    \begin{tabular}{lccccc}
      Scene                                                                                                       & GT                                                                                                  & Ours$_{\text{T}}$                                                                                      & Ours$_{\text{RGB}}$                                                                                    & GT                                                                                                      & Ours$_{\text{RGB}}$ \\
      Building                                                                                                    & \adjustbox{valign=c}{\includegraphics[width=.1\textwidth]{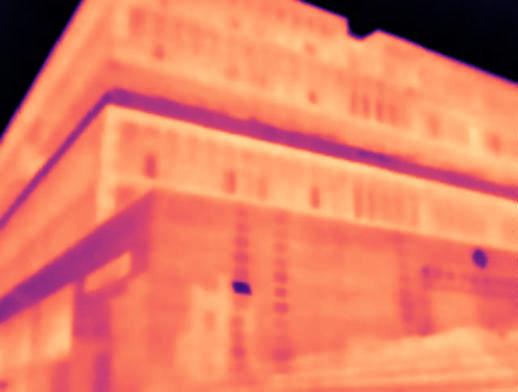}}     & \adjustbox{valign=c}{\includegraphics[width=.1\textwidth]{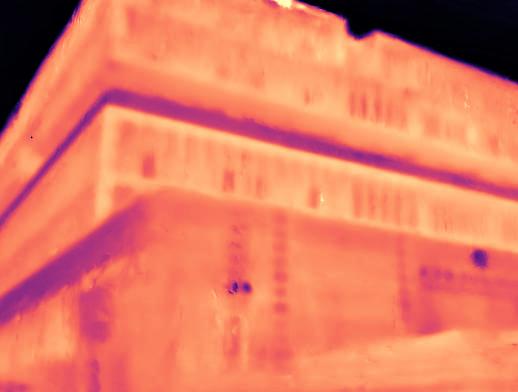}}     & \adjustbox{valign=c}{\includegraphics[width=.1\textwidth]{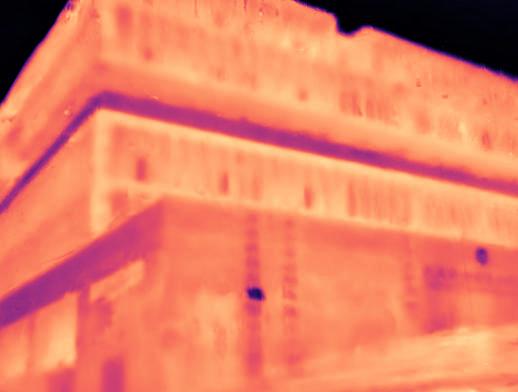}}     & \adjustbox{valign=c}{\includegraphics[width=.1\textwidth]{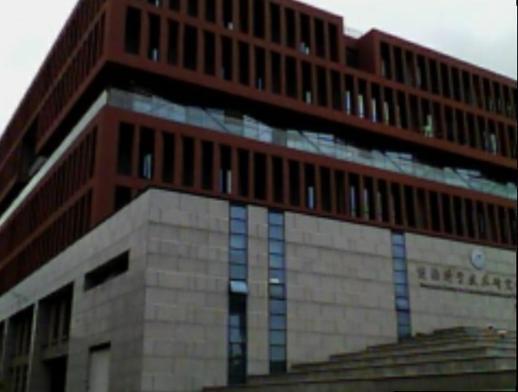}}     &
      \adjustbox{valign=c}{\includegraphics[width=.1\textwidth]{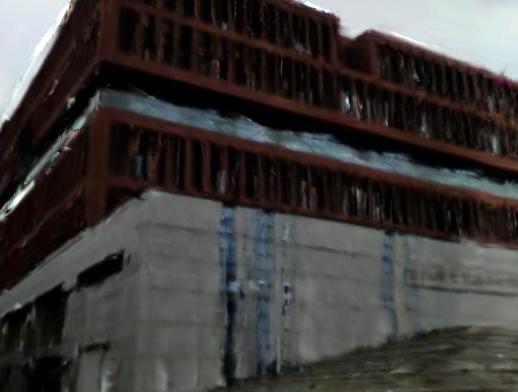}}                                                                                                                                                                                                                                                                                                                                                                                                                                                                    \\[1cm]

      \makecell[l]{Building A                                                                                                                                                                                                                                                                                                                                                                                                                                                                                                                                             \\(Spring)} & \adjustbox{valign=c}{\includegraphics[width=.1\textwidth]{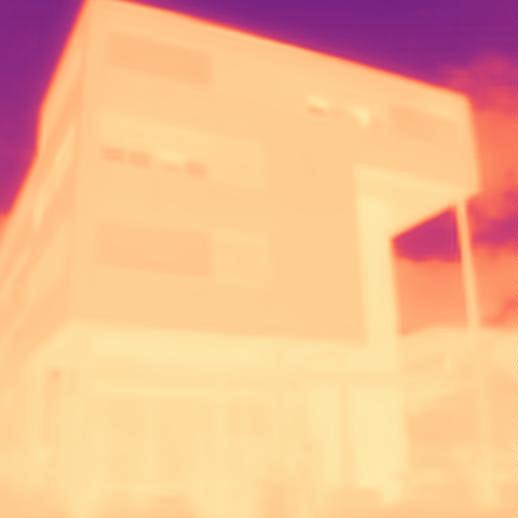}} & \adjustbox{valign=c}{\includegraphics[width=.1\textwidth]{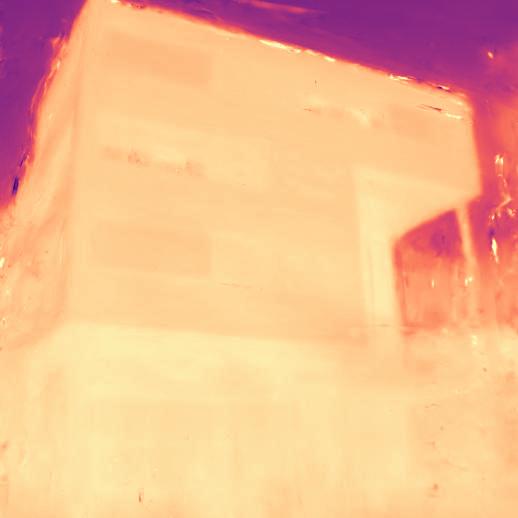}} & \adjustbox{valign=c}{\includegraphics[width=.1\textwidth]{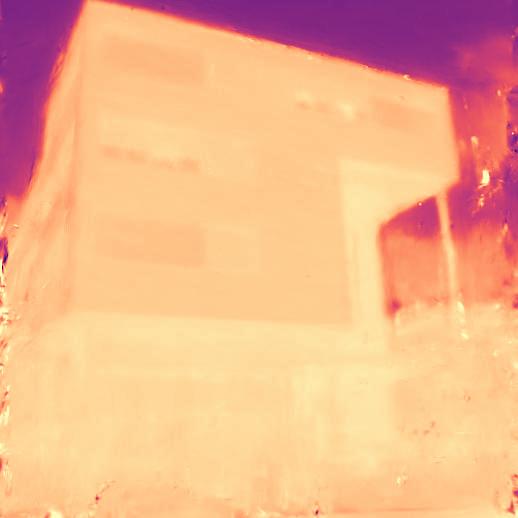}} &
      \adjustbox{valign=c}{\includegraphics[width=.1\textwidth]{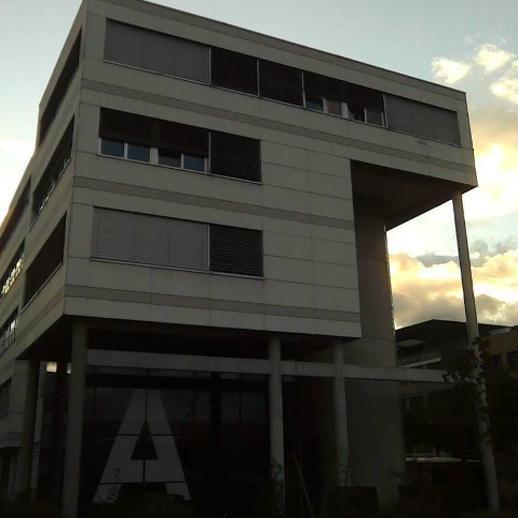}} &
      \adjustbox{valign=c}{\includegraphics[width=.1\textwidth]{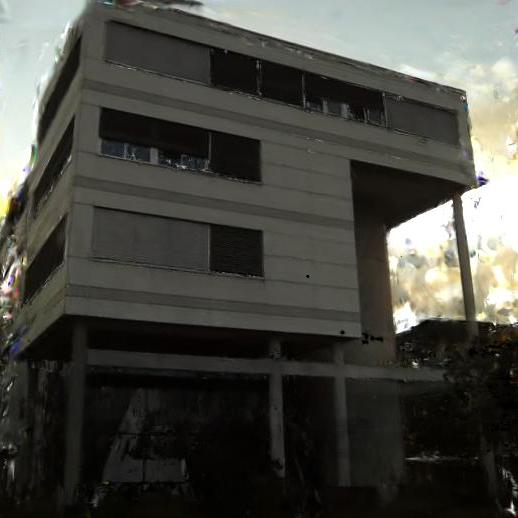}}                                                                                                                                                                                                                                                                                                                                                                                                                                                            \\[1cm]

      Kettle                                                                                                      & \adjustbox{valign=c}{\includegraphics[width=.1\textwidth]{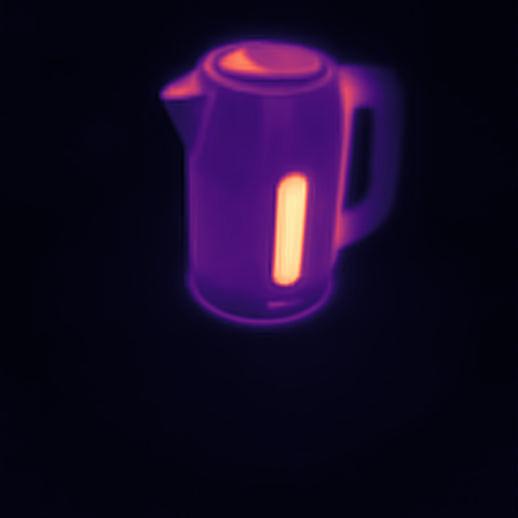}}       & \adjustbox{valign=c}{\includegraphics[width=.1\textwidth]{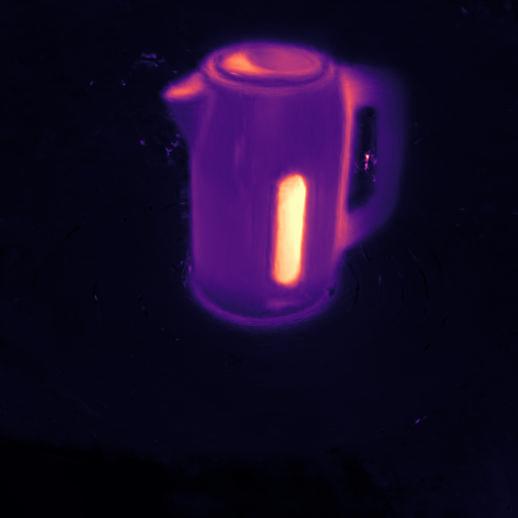}}       & \adjustbox{valign=c}{\includegraphics[width=.1\textwidth]{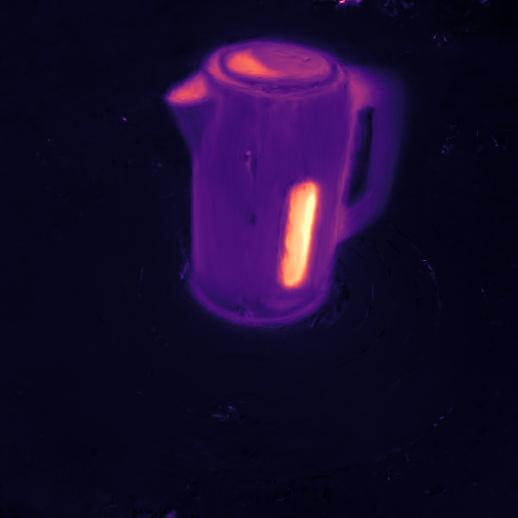}}       &
      \adjustbox{valign=c}{\includegraphics[width=.1\textwidth]{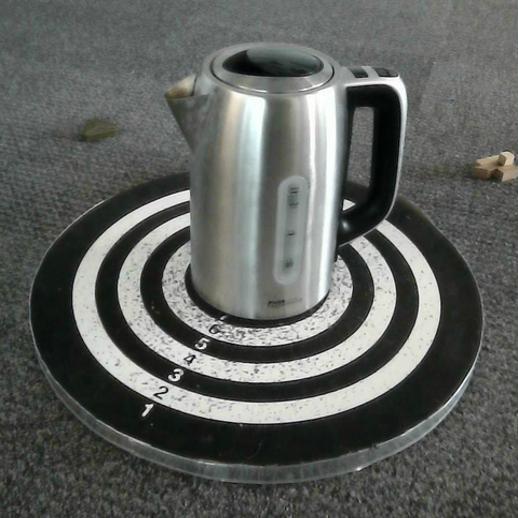}}           &
      \adjustbox{valign=c}{\includegraphics[width=.1\textwidth]{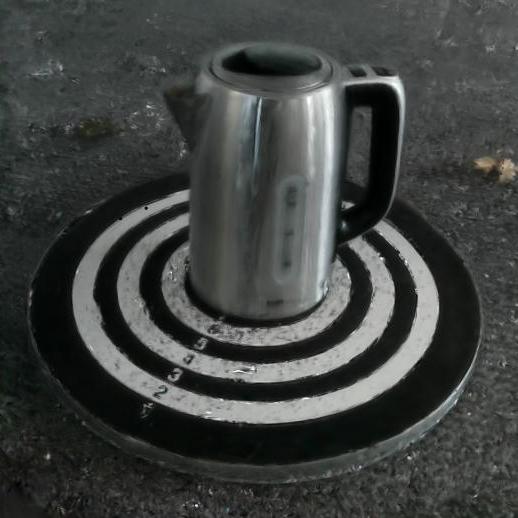}}                                                                                                                                                                                                                                                                                                                                                                                                                                                                      \\[1cm]

      Truck                                                                                                       & \adjustbox{valign=c}{\includegraphics[width=.1\textwidth]{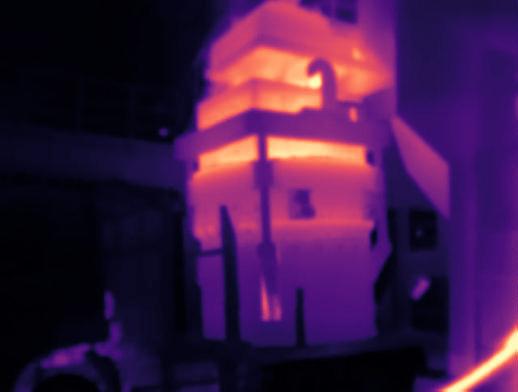}}        & \adjustbox{valign=c}{\includegraphics[width=.1\textwidth]{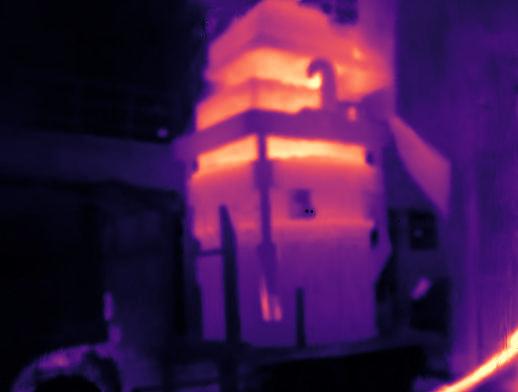}}        & \adjustbox{valign=c}{\includegraphics[width=.1\textwidth]{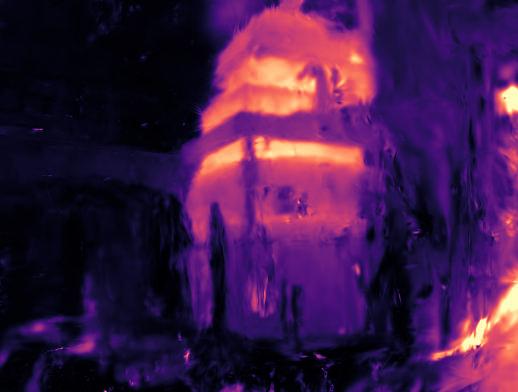}}        & \adjustbox{valign=c}{\includegraphics[width=.1\textwidth]{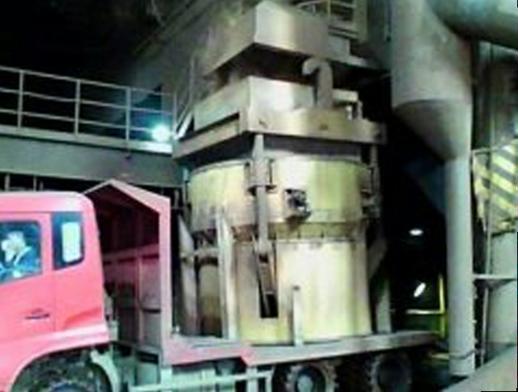}}        &
      \adjustbox{valign=c}{\includegraphics[width=.1\textwidth]{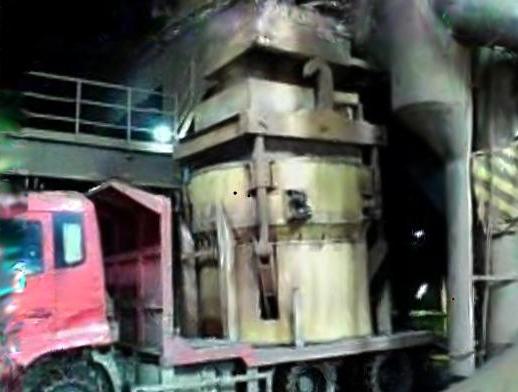}}                                                                                                                                                                                                                                                                                                                                                                                                                                                                       \\[1cm]

      INR Building                                                                                                & \adjustbox{valign=c}{\includegraphics[width=.1\textwidth]{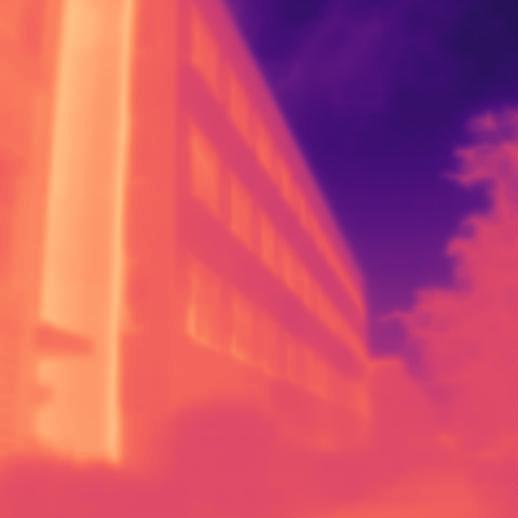}} & \adjustbox{valign=c}{\includegraphics[width=.1\textwidth]{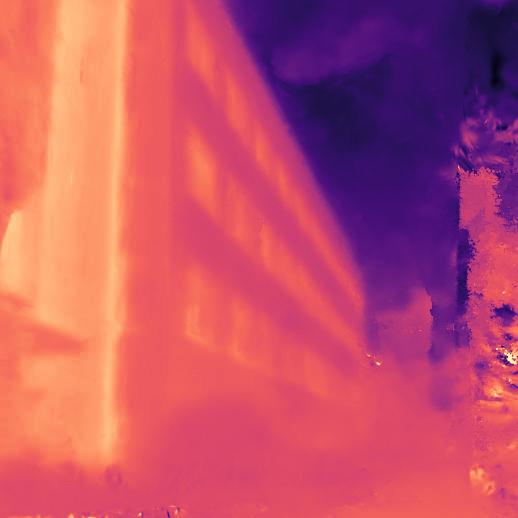}} & \adjustbox{valign=c}{\includegraphics[width=.1\textwidth]{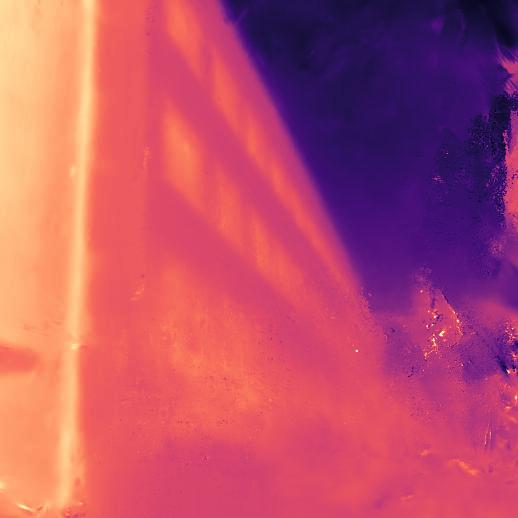}} & \adjustbox{valign=c}{\includegraphics[width=.1\textwidth]{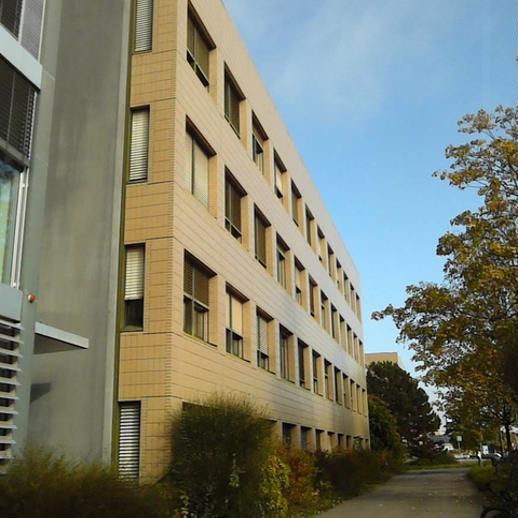}} &
      \adjustbox{valign=c}{\includegraphics[width=.1\textwidth]{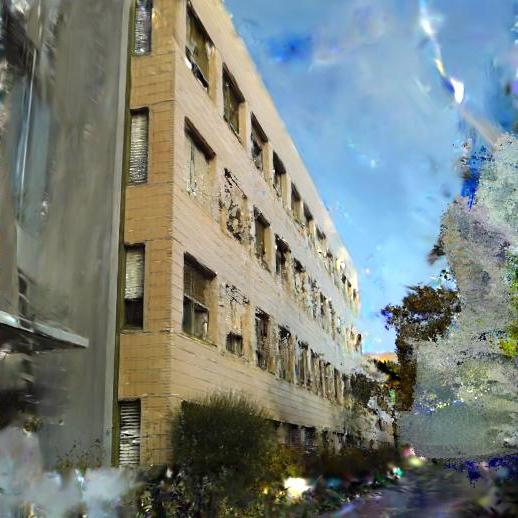}}                                                                                                                                                                                                                                                                                                                                                                                                                                                                \\

    \end{tabular}}

\end{table*}

\section{Implementation}

\subsection{Parameters}

We empirically set minimum number of XoFTR correspondences as $150$ of unimodal image pairs and $12$ for intermodal image pairs.
Our GD implementation uses the same default parameters as nerfacto.

\subsection{Dataset}

The aim of our paper is to show the challenges of benchmarking and developing multimodal 3D reconstructions.
To do so, we present a method that relies on existing feature detection and matching (i.e., XoFTR).
Thus, we conduct evaluations across nine diverse scenes sourced from ThermalGaussian~\cite{lu2025thermalgaussian} and ThermoScenes~\cite{hassanThermoNeRFMultimodalNeural2025}, selected based on a qualitative analysis of feature correspondences identified by XoFTR.
We excluded scenes where XoFTR failed to establish reliable correspondences, ensuring a fair evaluation of our core contributions, and left the exploration of other cross-modal feature extraction methods to future work.
For a detailed visualization of all scenes, refer to \cref{fig:eval:dataset}.

All scenes, whether from ThermoScenes or ThermalGaussian, include paired RGB and thermal images.
However, since our method is designed for unpaired data, we construct an unpaired dataset by: 1) randomly sampling either the RGB or thermal image from each pair, and 2) ensuring an equal count of RGB and thermal images.
This ensures no cross-modal correspondence remains.

\subsection{Metrics}

\textbf{Image Quality Metrics}
We evaluate both thermal and RGB modalities using standard quality metrics: Peak Signal-to-Noise Ratio (PSNR) and Structural Similarity Index (SSIM).
For RGB images, we additionally provide the Learned Perceptual Image Patch Similarity (LPIPS), as it aligns with human perceptual similarity---a property less relevant to thermal data.

\textbf{Temperature Fidelity Metrics}
To assess the accuracy of synthesized thermal images, we compute the Mean Absolute Error (MAE) between predicted and ground-truth temperature values.

\textbf{Camera Pose Evaluation}
Pose estimation is quantified using the Relative Pose Error (RPE) and the Absolute Trajectory Error (ATE) between thermal camera poses and equivalent RGB poses obtained via VGGT applied to \emph{the full} RGB sequence.
Given the lack of scale consistency between the nine scenes' reconstructions, all scenes were scaled to fit in a $1$m sphere before calculating the RPE and ATE to make the results comparable with each other.

\subsection{Inter VS Intra Modality Metrics}
\label{sec:eval:interintra}

In prior work~\cite{hassanThermoNeRFMultimodalNeural2025,tancik2023nerfstudio}, reconstruction is evaluated on the test set using SfM-derived poses (e.g., from COLMAP).
This is possible when either a single modality is used~\cite{tancik2023nerfstudio} or both modalities are paired and the RGB poses are applied to thermal images~\cite{hassanThermoNeRFMultimodalNeural2025}.
However, our approach requires independent pose estimation for each modality, and we have two independent sets of poses for the RGB and thermal test sets.

Hence, which reference pose (from RGB or thermal) to use to evaluate the quality of the novel views is an important question.
Naively adopting poses estimated using thermal images for thermal images and RGB-estimated poses for RGB images introduces a fundamental limitation: the absence of cross-modal geometric constraints in the scene representation.
Without such constraints, the method may optimize two independent scenes---one per modality---yielding deceptively high metrics while failing to produce a coherent multi-modal reconstruction; \cref{fig:eval:reference} illustrates this discrepancy by showing two novel-view renderings in RGB and thermal, demonstrating that the RGB and thermal reconstructions exhibit misalignment despite strong per-modality performance.

Thus, to evaluate our method for NVS, we measure both the inter-modal metrics, as well as the intra-modal metrics:

\subsubsection{Inter-Modal} RGB test images' camera parameters are assigned to the thermal evaluation images' parameters.
Hence, thermal evaluation images are the same as the RGB poses, and the Inter-Modal metrics allow for measuring that the thermal NVS is aligned with the RGB reconstruction.
It is important to note that, thus, a limitation of all unpaired multi-modal reconstruction frameworks is the need to have known poses for both modalities in the same reference frame, either through using paired images or using an external pose measurement setup---e.g., an IR pose tracker.
In our work, only in the test set, we use paired RGB and thermal images for which we obtain the RGB pose from VGGT and assign the corresponding RGB pose to the thermal images.
\subsubsection{Intra-Modal} RGB and thermal evaluation images use their own corresponding camera parameters, without taking into account the other modality.
In our work, those poses are obtained with VGGT.
With camera parameters of the evaluation images of both modalities independent, the \textit{Intra-Modal} metrics allow for measuring the consistency of each representation independently of the alignment between them.

We thus provide the intra-modal RGB and thermal metrics, as well as the inter-modal RGB (where RGB synthesized images are reconstructed from thermal poses) and inter-modal thermal (where thermal synthesized images are reconstructed from RGB poses)

\section{Experiments and discussion}

This section provides the evaluation results of our method against the closest method to ours, ThermalGaussian~\cite{lu2025thermalgaussian} (T-GS).
T-GS$_{\text{RGB}}$ corresponds to a model trained using RGB poses for both RGB and thermal images, while T-GS$_{\text{T}}$ corresponds to the same model trained with thermal poses.

On the other hand, our method is trained with the RGB and thermal camera poses found using the method presented in \cref{sec:method}.
The difference between ours$_{\text{RGB}}$ and ours$_{\text{T}}$ is not in the training---as it is for Thermal Gaussian---but only in the camera poses used for novel view synthesis.
Ours$_{\text{RGB}}$ corresponds to our results when the RGB poses are used for both RGB and thermal images synthesis from the test set, while ours$_{\text{T}}$ uses the thermal camera poses for both modalities.
Thus, looking at our definition of inter- and intra-modal metrics in \cref{sec:eval:interintra}, the intra-modal metrics correspond to the RGB results of ours$_{\text{RGB}}$ and the thermal results of ours$_{\text{T}}$, while the inter-modal metrics correspond to the RGB results of ours$_{\text{T}}$ and the thermal results of ours$_{\text{RGB}}$.

While this comparison is not strictly fair---since our method assumes no prior modality alignment, whereas the baseline operates under ideal, fully paired conditions---it provides a quantitative upper bound, revealing how closely our unpaired reconstruction approaches the ideal scenario.

\subsection{Novel View Synthesis}

Looking at \cref{tab:per-scene-metrics-all}, one can see the impact of using inter- versus intra-modal metrics.
One can see a decrease in the metrics of the thermal images synthesis between ours$_{\text{RGB}}$ and ours$_{\text{T}}$, where thermal images evaluated against the thermal poses (i.e., ours$_{\text{T}}$) have higher quality than when synthesized from their equivalent RGB pose (i.e., ours$_{\text{RGB}}$).
For example, the PSNR decreases from $25.595$ to $21.315$, the SSIM from $0.851$ to $0.771$, and the MAE increases from $2.509$ to $4.914$.
On the other hand, the quality of image synthesis in RGB remains stable regardless of the reference used, with only a small decrease from RGB to thermal.

Interestingly, when comparing to T-GS, our method performs better when performing image synthesis from thermal camera poses for both RGB and Thermal images (ours$_{\text{T}}$ versus T-GS$_{\text{T}}$).
On the other hand, from RGB poses (ours$_{\text{RGB}}$ versus T-GS$_{\text{RGB}}$), we obtain lower results.
Those are expected results since VGGT-estimated poses from RGB images are more accurate than thermal images; hence, using only thermal camera poses for training performed worse than our mix of RGB and thermal poses, which in turn performed worse than only using RGB poses for all images.

\subsection{Pose Estimation}

To evaluate the quality of the pose alignment between thermal and RGB, we align the thermal poses via the Procrustes transform computed as in \cref{sec:method:procrustes} on the half-datasets.
Then, we estimate the poses for the full RGB dataset using VGGT and evaluate the Absolute Trajectory Error (ATE) and Relative Pose Error (RPE) between the thermal camera poses and the corresponding RGB poses from the full dataset.
Since VGGT does not guarantee scale between reconstructed scenes, each scene is scaled to fit in a 1m unitsphere before hand.

Results, presented in \cref{tab:eval:metrics}, show the RMSE, mean, and median ATE and RPE over all scenes.
One can see that, while the metrics are relatively low (for example, the highest ATE RMSE in position only $0.158$ for the Building scene), there is still a difference between RGB and thermal poses.
Hence, the alignment is not perfect; it should be taken into account that, since the camera parameters are estimated by VGGT (as explained in \cref{sec:method:vggt}), the difference could be due to inaccuracy in the camera parameter estimation.
For example, an inaccurate focal lens can lead to misaligned thermal and RGB camera poses, but accurate scene reconstruction and image synthesis.

\begin{table}[t]
    \vspace{1mm}
    \centering
    \caption{
        Average of the mean, median, and RMSE for position and rotation over the 9 test scenes.
    }
    \label{tab:eval:metrics}
    \begin{tabular}{lcc}
        \toprule
        Metric      & mean [m/m]      & RMSE [m/m]       \\ \midrule
        RPE$_{pos}$ & $0.10 \pm 0.06$ & $0.13 \pm 0.08$  \\
        RPE$_{rot}$ & $3.43 \pm 1.45$ & $4.12 \pm 1.89$  \\
        ATE$_{pos}$ & $0.08 \pm 0.06$ & $0.085 \pm 0.05$ \\
        ATE$_{rot}$ & $0.07 \pm 0.04$ & $0.075 \pm 0.04$ \\

        \bottomrule
    \end{tabular}
\end{table}

\subsection{Manual scale invariant alignment}

To showcase how 3D multi-modal GS representations are able to learn two independent RGB and Thermal representations when poses are not well aligned, we provide a study to compare our results to reconstructions where the poses were manually aligned using a known initial pose and ICP.

\begin{figure}[t]
    \vspace{2mm}
    \centering
    \begin{subfigure}[t]{0.1\textwidth}
        \centering
        \includegraphics[width=\textwidth]{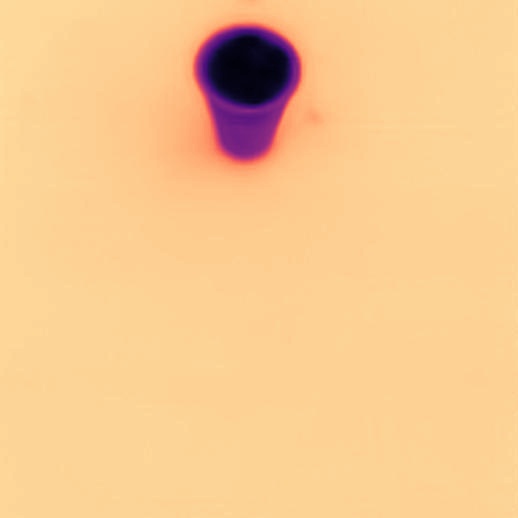}
        \label{fig:sub1}
    \end{subfigure}
    \hfill
    \begin{subfigure}[t]{0.1\textwidth}
        \centering
        \includegraphics[width=\textwidth]{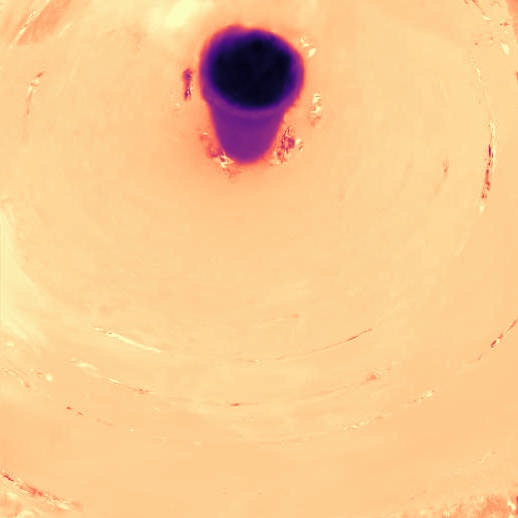}
        \label{fig:sub2}
    \end{subfigure}
    \hfill
    \begin{subfigure}[t]{0.1\textwidth}
        \centering
        \includegraphics[width=\textwidth]{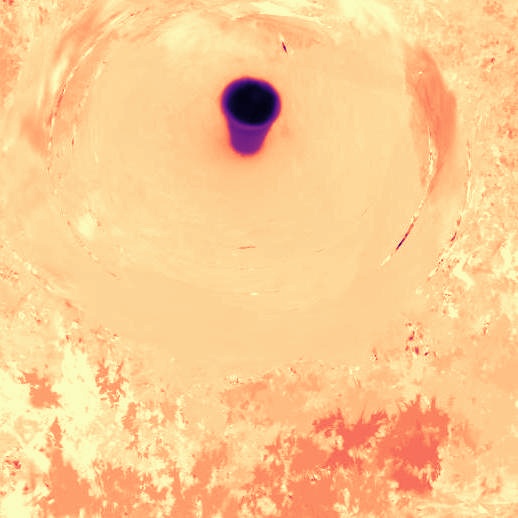}
        \label{fig:sub3}
    \end{subfigure}
    \hfill
    \begin{subfigure}[t]{0.1\textwidth}
        \centering
        \includegraphics[width=\textwidth]{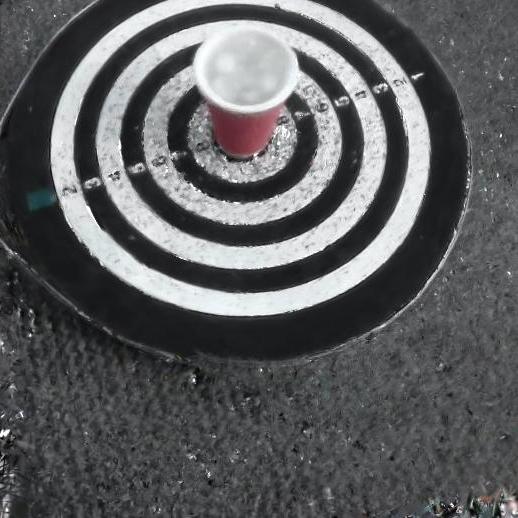}
        \label{fig:sub4}
    \end{subfigure}
    \\
    \begin{subfigure}[t]{0.1\textwidth}
        \centering
        \includegraphics[width=\textwidth]{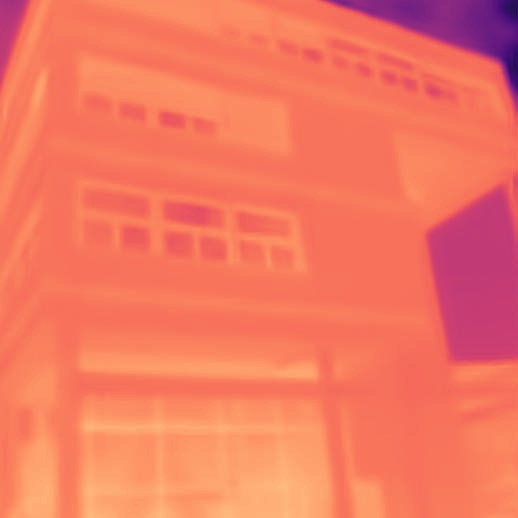}
        \label{fig:sub5}
    \end{subfigure}
    \hfill
    \begin{subfigure}[t]{0.1\textwidth}
        \centering
        \includegraphics[width=\textwidth]{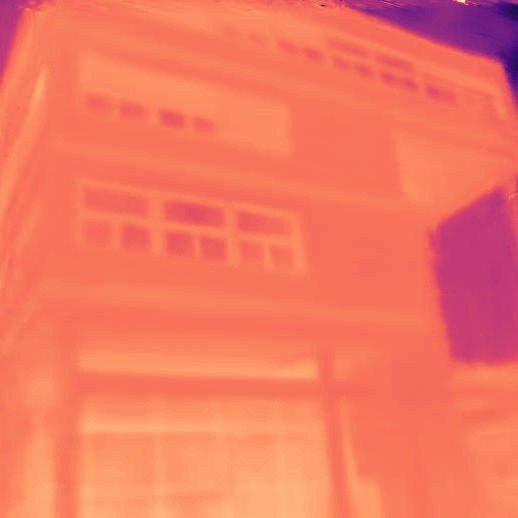}
        \label{fig:sub6}
    \end{subfigure}
    \hfill
    \begin{subfigure}[t]{0.1\textwidth}
        \centering
        \includegraphics[width=\textwidth]{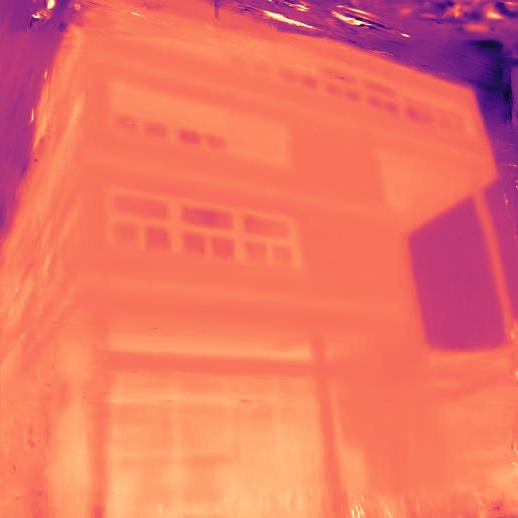}
        \label{fig:sub7}
    \end{subfigure}
    \hfill
    \begin{subfigure}[t]{0.1\textwidth}
        \centering
        \includegraphics[width=\textwidth]{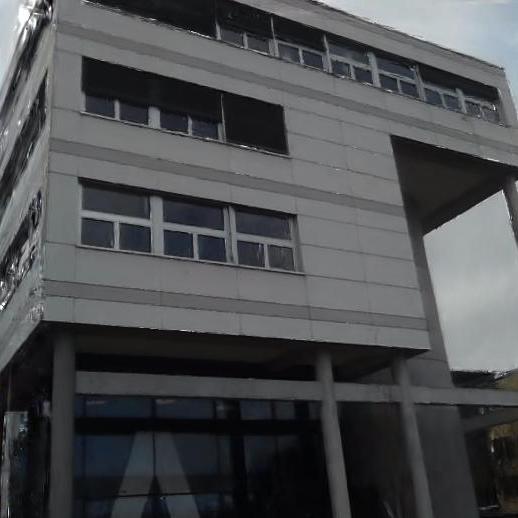}
        \label{fig:sub8}
    \end{subfigure}
    \caption{Using a reference image to align the camera poses instead of our method, one can see a shift between the reconstructed thermal images from RGB poses (third column) and the one reconstructed from the RGB pose (fourth column).
        On the other hand, looking at the thermal image synthesis from thermal camera poses (second column), the images are correct: the model has learn two different 3D representations for each modality instead of a multi-modal representation.
    }
    \vspace{-6mm}
    \label{fig:eval:reference}
\end{figure}

Using this strategy, the camera poses between modalities are simply aligned based on a known reference.
Given that VGGT doesn't guarantee a scale between each reconstruction (RGB and thermal), we use ICP to further align the camera poses without changing the scale.
Image synthesis for two scenes can be seen in \cref{fig:eval:reference}.
One can observe that, while the image synthesis from each modality's camera poses (i.e., from RGB camera poses from RGB, and thermal camera poses for thermal) is accurate, if we assume that equivalent RGB/thermal camera poses should return similar images, the results are incorrect.
There is a large gap between the second and third column of \cref{fig:eval:reference}, and between the third and fourth.

It shows that pose alignment between modalities is essential to build a truly multi-modal 3D representation, since otherwise, two independent representations, one for each modality, can be built.
This also highlights the importance of using both inter- and intra-modal metrics without relying only on intra-modal metrics.

\section{Conclusion and Future Work}

This work enables multi-modal RGB-thermal novel view synthesis (NVS) from unpaired image sets, eliminating the need for paired data or stereo setups.
By combining VGGT for independent RGB and thermal camera parameter estimation and XoFTR for feature-based alignment, we achieve cross-modal alignment between unpaired RGB and thermal images.
Moreover, we propose a 3D Gaussian Splatting-based training strategy for unpaired input data.

We also highlight the limitations of current evaluation strategies for multi-model 3D reconstruction and image synthesis and provide a new strategy that accounts for both intra-modal (per-modality consistency) and inter-modal (cross-modal coherence) metrics.
Indeed, our experiments reveal that alignment is critical for true multi-modal representation: Without proper pose registration, models may learn separate RGB and thermal geometries, yielding high intra-modal metrics but failing to capture cross-modal relationships.

Future work will explore more robust cross-modal matching (e.g., for low-texture or extreme thermal variations).
By eliminating the need for paired data, this framework paves the way for scalable, practical multi-modal 3D reconstruction.

\printbibliography

\end{document}